\documentclass[journal]{IEEEtran}

\usepackage{cite}
\usepackage[caption=false,font=footnotesize]{subfig}
\usepackage{latexsym}
\usepackage{etoolbox}
\usepackage{url}
\usepackage{booktabs}
\usepackage{algorithmic}
\usepackage{array}
\usepackage{graphicx}
\usepackage{booktabs}
\usepackage[flushleft]{threeparttable}
\usepackage{amsmath}
\usepackage{textcomp}
\usepackage{xspace}
\usepackage{xcolor}
\usepackage[version-1-compatibility]{siunitx}
\usepackage{multirow}

\newcommand{\DAVIDE}{D.A.V.I.D.E.\xspace}
\newcommand{\DIG}{DiG\xspace}
\newcommand{\pae}{pAElla\xspace}
\newcommand{\quotes}[1]{``#1''}

\newcommand{\eg}{\textit{e.g.},\xspace}
\newcommand{\ie}{\textit{i.e.},\xspace}
\newcommand{\fscore}{F1-score\xspace}
\newcommand{\IoT}{Internet-of-Things\xspace}
\newcommand{\DC}{DC\xspace}
\newcommand{\DCs}{DCs\xspace}
\newcommand{\SC}{SC\xspace}
\newcommand{\SCs}{SCs\xspace}
\newcommand{\DCSC}{DC\,/\,SC\xspace}
\newcommand{\DCSCs}{DCs\,/\,SCs\xspace}
\newcommand{\aka}{a.k.a.\xspace}
\newcommand{\wrt}{w.r.t.\xspace}

\begin{document}

\title{\textit{\pae}: Edge-AI based Real-Time Malware Detection in Data Centers}

\author{Antonio~Libri,~\IEEEmembership{Member,~IEEE,}
        Andrea~Bartolini,~\IEEEmembership{Member,~IEEE,}
        and~Luca~Benini,~\IEEEmembership{Fellow,~IEEE}
\thanks{A. Libri and L. Benini are with D-ITET, ETH Zurich, Zurich, Switzerland (e-mail:~\{a.libri,lbenini\}@iis.ee.ethz.ch).}
\thanks{A. Bartolini is with DEI, University of Bologna, Bologna, Italy (e-mail:~a.bartolini@unibo.it).}
\thanks{L. Benini is also with DEI, University of Bologna, Bologna, Italy (e-mail:~luca.benini@unibo.it).}
}


\maketitle


\begin{abstract}
The increasing use of \IoT (IoT) devices for monitoring a wide spectrum of applications, along with the challenges of \quotes{big data} streaming support they often require for data analysis, is nowadays pushing for an increased attention to the emerging edge computing paradigm. In particular, smart approaches to manage and analyze data directly on the network edge, are more and more investigated, and Artificial Intelligence (AI) powered edge computing is envisaged to be a promising direction. In this paper, we focus on Data Centers (\DCs) and Supercomputers (\SCs), where a new generation of high-resolution monitoring systems is being deployed, opening new opportunities for analysis like anomaly detection and security, but introducing new challenges for handling the vast amount of data it produces. In detail, we report on a novel lightweight and scalable approach to increase the security of \DCSCs, that involves AI-powered edge computing on high-resolution power consumption. The method - called \pae~- targets real-time Malware Detection (MD), it runs on an out-of-band IoT-based monitoring system for \DCSCs, and involves Power Spectral Density of power measurements, along with AutoEncoders. Results are promising, with an F1-score close to 1, and a False Alarm and Malware Miss rate close to 0\%. We compare our method with State-of-the-Art MD techniques and show that, in the context of \DCSCs, \pae can cover a wider range of malware, significantly outperforming SoA approaches in terms of accuracy. Moreover, we propose a methodology for online training suitable for \DCSCs in production, and release open dataset and code.
\end{abstract}

\begin{IEEEkeywords}
Artificial Intelligence, Edge Computing, Malware Detection, IoT monitoring, Data Center, Supercomputer.
\end{IEEEkeywords}

%
\IEEEpeerreviewmaketitle

\section{Introduction}

\IEEEPARstart{I}{n} the era of the Internet-of-Things (IoT), commodity devices such as low-cost embedded systems can easily compose sensor networks and are increasingly used as monitoring infrastructures, from household appliances to industrial environments~\cite{Samuel16,Pan_2018,Zhang_2016,SmarTEG19}. Depending on the target application, they can produce a massive amount of data that needs to be handled, and to tackle this challenge the emerging edge computing paradigm is receiving a tremendous amount of interest~\cite{Chen_1,Chen_2,Yang18}. Indeed, by pushing data storage and analysis closer to the network edge, this approach allows to mitigate the network traffic load and meet requirements such as high scalability, low latency, and real-time response. However, at its basis, smart techniques to manage and analyze data directly on edge are of crucial importance. With this motivation, emerging methods based on Artificial Intelligence (AI) running on edge devices (\aka edge AI or edge intelligence) are envisioned as a promising solution to address this challenge~\cite{Chen_3,Chen_4}.

This is true also in the context of Supercomputers (\SCs) and Data Centers (\DCs), where a new generation of high-resolution monitoring systems is being deployed~\cite{DiG_DAAC18,DiG_Extended,HAEC}, pushing the research boundaries over new opportunities to support their automation, analytics, and control. In particular, \DIG~\cite{DiG_DAAC18} is the first out-of-band monitoring system for \DCSCs that allows real-time edge intelligence on high-resolution measurements, and is already installed in a \SC in production (\ie \DAVIDE, ranked 18\textsuperscript{th} most efficient \SC in the world, based on Green500 November 2017~\cite{DAVIDE,BigDaw,BartPav19}). \DIG is based on low-cost IoT embedded computers, which are essential to interface with analog power sensors to collect very fine-grain power measurements of the \DC servers (\ie up to \SI{20}{\micro\s}). Moreover, the system provides real-time on-board processing capabilities, coupled with ultra-precise time stamping (\ie below microseconds) and a vertical integration in a scalable distributed data analytics framework (\ie ExaMon~\cite{ExaMon}). 

\begin{figure}[ht]
\centering
\includegraphics[width=3.4in]{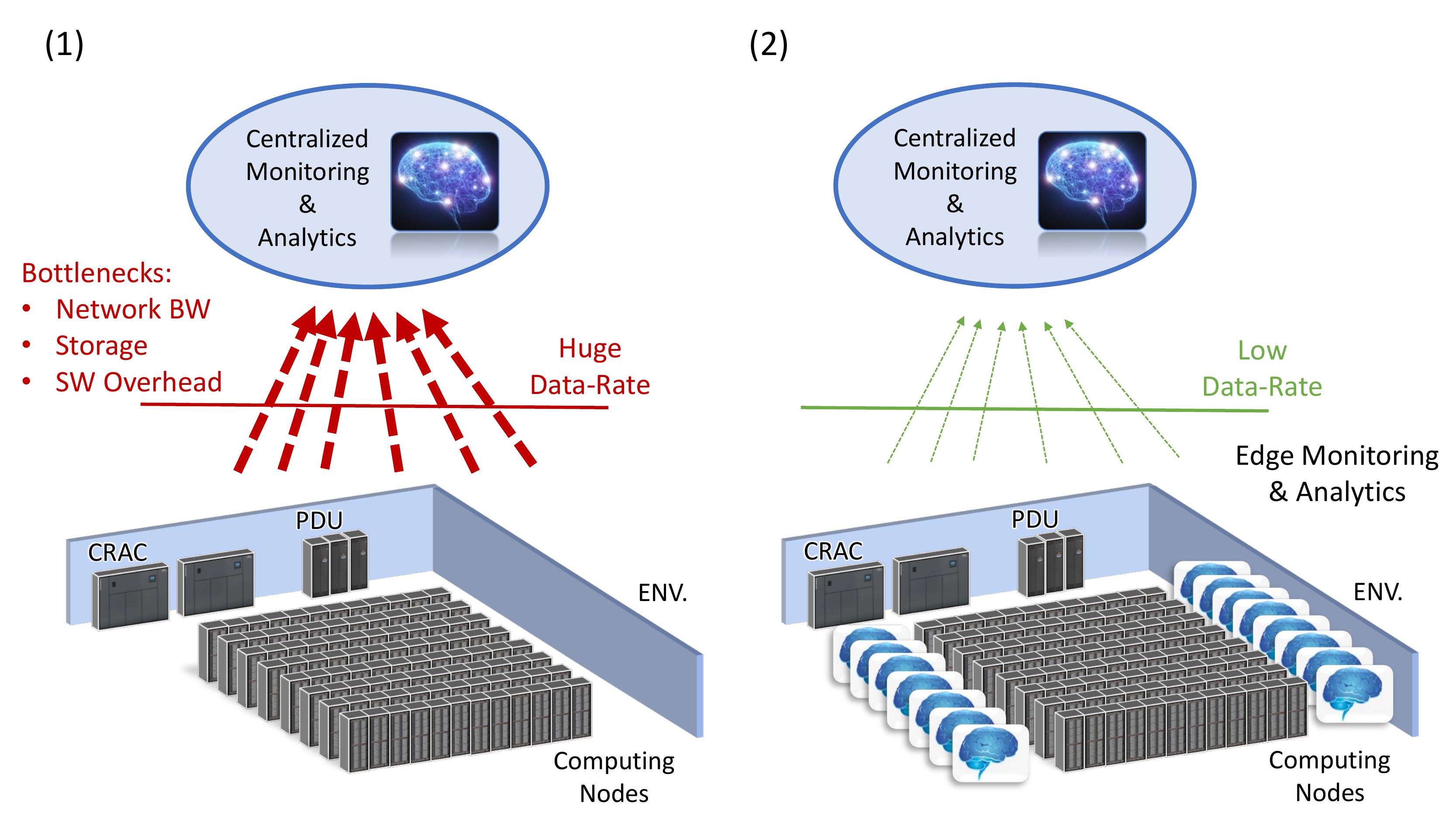}
\caption{High-Resolution monitoring bottlenecks in \DCs.}
\label{fig:dc}
\end{figure}

These unique features allow us to carry out real-time high-frequency analysis that would not be possible otherwise: with in-band monitoring, like Intel RAPL, the time granularity is limited to the order of millisecond~\cite{RAPL_intel}, while with current out-of-band monitoring solutions it is not possible to carry out on-board edge analytics~\cite{DiG_DAAC18}. As depicted in Figure~\ref{fig:dc}.1, using an out-of-band monitoring with high-resolution measurements (\ie below millisecond) and no edge computing capabilities, would result in crucial bottlenecks, such as (i) overhead on the network bandwidth, (ii) overhead on the data storage capacity (to save measurements for post-processing analysis), (iii) and overhead on the software tools that have to handle the measurements (in real-time and offline)~\cite{DiG_DAAC18}. As matter of example, supposing to install a high-resolution power monitoring system like HAEC~\cite{HAEC}, with a sampling rate of \SI{500}{\kilo S/\s} (kilo Samples per second) on 4 custom sensors, on a large-scale \SC like \textit{Sunway TaihuLight} (3\textsuperscript{rd} in Top500 of June 2019, and involving around 41 thousand compute nodes~\cite{Fu2016}), it would require a data collection network bandwidth of \texttildelow\SI{82}{\giga S/\s}, with obvious overheads on software and storage to handle it. Instead, with a \DIG-like monitoring infrastructure (Figure~\ref{fig:dc}.2) we can exploit distributed computing resources to carry out dedicated analysis for each server (server-level analytics), and send at a much lower rate the detected events to a centralized monitoring unit, which has the complete view and holistic knowledge of the whole cluster (cluster-level analytics).

However, to perform real-time high-frequency analysis at the edge, smart data science approaches are essential. In this paper, we show how a novel approach involving AI and Machine Learning (ML) running on IoT monitoring devices, can allow increasing the cybersecurity of \DCs, which is nowadays a high impact research topic~\cite{Jensen09,Watson16,Popovic10}. Our edge AI approach targets real-time Malware Detection (MD) for \DCSCs, and involves feature extraction analysis based on Power Spectral Density (PSD) estimation with the Welch method, along with an AutoEncoder (AE) - which is a particular Neural Network (NN) suitable for Anomaly Detection (AD) - to identify patterns of malware. While similar approaches are already used in the context of speech recognition~\cite{deng2010,wan2017google}, to the best of our knowledge, this is the first time is used together with high-resolution power measurements for malware - and more in general anomaly - detection. We called the approach \textit{\pae}, which stands for \quotes{Power-AutoEncoder-WeLch for anomaLy and Attacks}.
\smallskip\smallskip

\textit{Contributions of the work:}
\begin{enumerate}
    \item A novel out-of-band approach, \pae, running on IoT-based monitoring systems for real-time MD in \DCs. Results in our big dataset (\ie more than ninety malware + 7 benchmarks, representative of \SC and \DC activity) report an overall F1-score close to 1, and a False Alarm and Malware Miss rate close to \SI{0}{\percent}. We compare our approach with other State-of-the-Art (SoA) ML techniques for MD and show that, in the context of \DCSCs, \pae can cover a wider range of malware, outperforming the best SoA method by \SI{24}{\percent} in terms of accuracy. In addition, notice that this approach can be used more in general also for AD, in cases where high resolution becomes essential to identify patterns, opening new opportunities for researchers.
    \item We propose a methodology for online training in the \DC infrastructure suitable to be run in a system in production like \DIG.
    \item The approach involves zero overhead on the servers processing elements, and it is suitable for large-scale systems, thanks to the on-board ML inference running at the edge, on each compute node.
    \item We release both dataset and software we used for the analysis~\cite{pae_dataset}. The dataset includes 7 benchmarks (6 scientific applications for \SCs + the signature of the system in idle) and 95 malware of different sort. Notice that this is the first dataset of its kind, providing to the security research community high-resolution measurements (\ie power measurements @\SI{20}{\micro\s} + performance measurements @\SI{20}{\milli\s}) for carrying out and benchmarking novel analyses.
\end{enumerate}

Finally, we highlight the fact that \pae is suitable to be used in a real \DCSC in production, such as in \DAVIDE, that integrates \DIG~\cite{DAVIDE}.
\smallskip

\textit{Outline:} Section~\ref{sec:malw} introduces background information on performance counters based MD. Section~\ref{sec:rw} reports the related work. Section~\ref{sec:dig} provides an overview of the \DIG infrastructure that we use to carry out our analysis. Section~\ref{sec:pAE} presents both the \pae algorithm and its implementation on \DIG, along with the methodology to run online training in a \DC. Finally, we show in Section~\ref{sec:res} how we built the dataset for our analysis, together with the results of the MD and a comparison with SoA.  We also report an analysis of the overheads in running \pae on \DIG, and some considerations on its scalability to large-scale \DCs. We conclude the paper in Section~\ref{sec:end}.

\section{Background on Malware Detection}\label{sec:malw}

The threat of malware and more in general cyber attacks is nowadays considerably increasing, and countermeasures to detect them are becoming more critical than ever. Examples spread from different kinds of backdoors, to trojans and ransomware, each of them having different Operating System (OS) and computational usage characteristics~\cite{tang2014}. A prominent SoA technique for MD - and more in general for AD - is based on catching their dynamic micro-architectural execution patterns, employing monitoring the hardware performance counters available in modern processors~\cite{Demme2013,tang2014,Sayadi2018,AysePerf2019,VM_Cloud2017}. By using this approach, we can benefit of (i) a lower overhead than using higher-level approaches - \eg monitoring patterns at application and OS level -, and (ii) also of the fact that these micro-architectural features are efficient to audit and harder for attackers to control, when they want to avoid detection in evasion attacks~\cite{tang2014}.

There are mainly two ways to train the anomaly-based detector to catch malware: (i) training on malware signatures~\cite{Demme2013}, or (ii) training on \quotes{healthy} operating conditions of the system that we want to protect by malware attacks, to be able to identify possible changes that indicate an infection~\cite{tang2014}. While the first method is a supervised learning approach, the second one can be seen as \quotes{semi-supervised} learning as we do not need any prior knowledge of the malware signatures, thus we can detect even zero-days attacks (\ie attacks which are never seen before)~\cite{BorghesiAD2019,tang2014}. The idea behind the second method can also be applied within the context of \DCs and \SCs: the supercomputing center can study the \quotes{regular} activity of their users, and thus create models on healthy machine states.

\textit{Drawbacks:} 
\begin{itemize}
    \item it is not possible to monitor all the performance counters at the same time (\eg Intel allows the selection of only 4 performance counter at a time via the Performance Monitoring Unit - PMU -, which can become 8 by disabling the Hyperthreading\cite{Intel_PMU_Manual}). This does not allow to be flexible, as some malware are better described by certain performance counters rather than others, with a consequent degradation on the malware accuracy detection (we do not know a priori which malware is running, so which performance counter is better to select).
    \item Depending on the running malware, the time granularity of the monitoring becomes of primary importance to be able to catch their micro-architectural execution patterns~\cite{tang2014}. However, as described in the Introduction Section, current SoA built-in tools to monitor performance counters are limited by sampling rate and edge computing capabilities.
\end{itemize}

As we show in our results, these drawbacks can be a limitation to discover malware in real-time in a \DC.
\smallskip

\textit{Our Approach:} To bridge this gap, we propose a solution that exploits the out-of-band and very high resolution power consumption measurements of \DIG (no performance counters involved), in a lightweight algorithm that can run in real-time on the IoT devices at the edge of the \DC. This method allows catching fine-grain activities of the malware that are not visible with the performance counter based approach.

\section{Related Work}\label{sec:rw}

\textit{Perf-Counters-based MD:} In recent years, several works in the literature focused on the usage of performance counters for Anomaly and Malware Detection. In particular, \cite{Demme2013} showed the feasibility of the technique by testing them with several supervised learning approaches (\ie k-Nearest Neighbors, decision tree, random forest, and Artificial Neural Networks) after training models on known malware. Later, \cite{tang2014} showed the feasibility of this method with an unsupervised learning approach, namely the one-class Support Vector Machine (oc-SVM), by training the model on healthy programs. Moreover, they suggested a list of performance counters that can, on average, better describe the malware signatures (we use this list - reported in Table~\ref{tab:perf_core} - to compare this method with our approach). In this direction, different methods were proposed to improve the technique, not only for MD, but also more in general for AD. In the context of \DCs, works in \cite{AyseTax2018,AysePerf2019,VM_Cloud2017,BorghesiAD2019} showed how to use performance counters to detect anomalies, while \cite{minigTahir17} showed how to use them for detecting covert cryptocurrency mining. However, very recent works in~\cite{Zhou2018,das2019sok} bring into question the robustness of this technique for security, carrying out a study with a big dataset with more than ninety malware and reporting poor results in accuracy detection.
\smallskip

\textit{Power-based MD:} Similar to the performance-counters-based approach, several works in literature proposed methods to detect malicious activity in IoT networks by mean of energy consumption footprint. To provide some examples, \cite{Azmoodeh2018} presented a machine learning based approach to detect ransomware attacks by monitoring the power consumption of Android devices, or \cite{Zhang14} proposed a cloud-assisted framework to defend connected vehicles against malware. While all these works targeted the security of embedded systems, a proof of concept toward the feasibility of power-based malware detection for general-purpose computers was proposed in~\cite{Bridges18}. In this paper, Bridges et al. use an unsupervised one-class anomaly detection ensemble, based on statistical features (\eg mean, variance, skewness and Kurtosis), with a power consumption sampling rate of \SI{17}{\milli\s}. However, as they claim in their paper, further research to increase the sampling rate is necessary, especially for accurate baselining of very small instructions sequences of malware. Moreover, although this was the first step on this kind of approach, to the best of our knowledge: (i) no other works proved yet the feasibility in a \DC where embedded monitoring networks are involved, and edge intelligence for real-time analytics is required; (ii) there is not yet a robust dataset with several malware that proves the robustness of the technique.
\smallskip

\textit{Data Center Monitoring:} Off-the-shelf methods to measure power and performance of \DC compute nodes rely on in-band or out-of-band telemetry depending on the technology vendors. In particular, an example of in-band solution is Intel RAPL~\cite{RAPL_intel}, which can provide a time granularity up to \SI{1}{\milli\s}, while examples of out-of-band solutions are IBM Amester~\cite{rosedahl2017} and IPMI~\cite{IPMI}, which can provide a resolutions of \SI{250}{\micro\s} and \SI{1}{\s}, respectively, but not AI-powered edge computing. Towards high-resolution monitoring of \DCs, several works in recent years proposed solutions, such as~\cite{HDEEM,HAEC}, and \DIG~\cite{DiG_DAAC18} is the best-in-class SoA infrastructure that bridges the gap of very fine-grain monitoring (\ie \SI{20}{\micro\s} of time granularity) and edge computing.
\smallskip

\textit{PSD and NNs:} While NNs, and more in particular AEs, are already used for anomaly and malware detection in different domains, including \DCs~\cite{BorghesiAD2019,Wang2019}, the use of Fourier analysis together with NNs is a well-known procedure in speech recognition~\cite{deng2010,wan2017google}. We use a similar approach for malware detection in our study, involving PSD with Welch method, and an AE for ML inference on-board. To the best of our knowledge, this is the first work proving its feasibility for the cybersecurity of \DCs.

\section{\DIG \& AD on the edge of a Top500's Supercomputer}\label{sec:dig}

\begin{figure}[ht]
\centering
\includegraphics[width=3.4in]{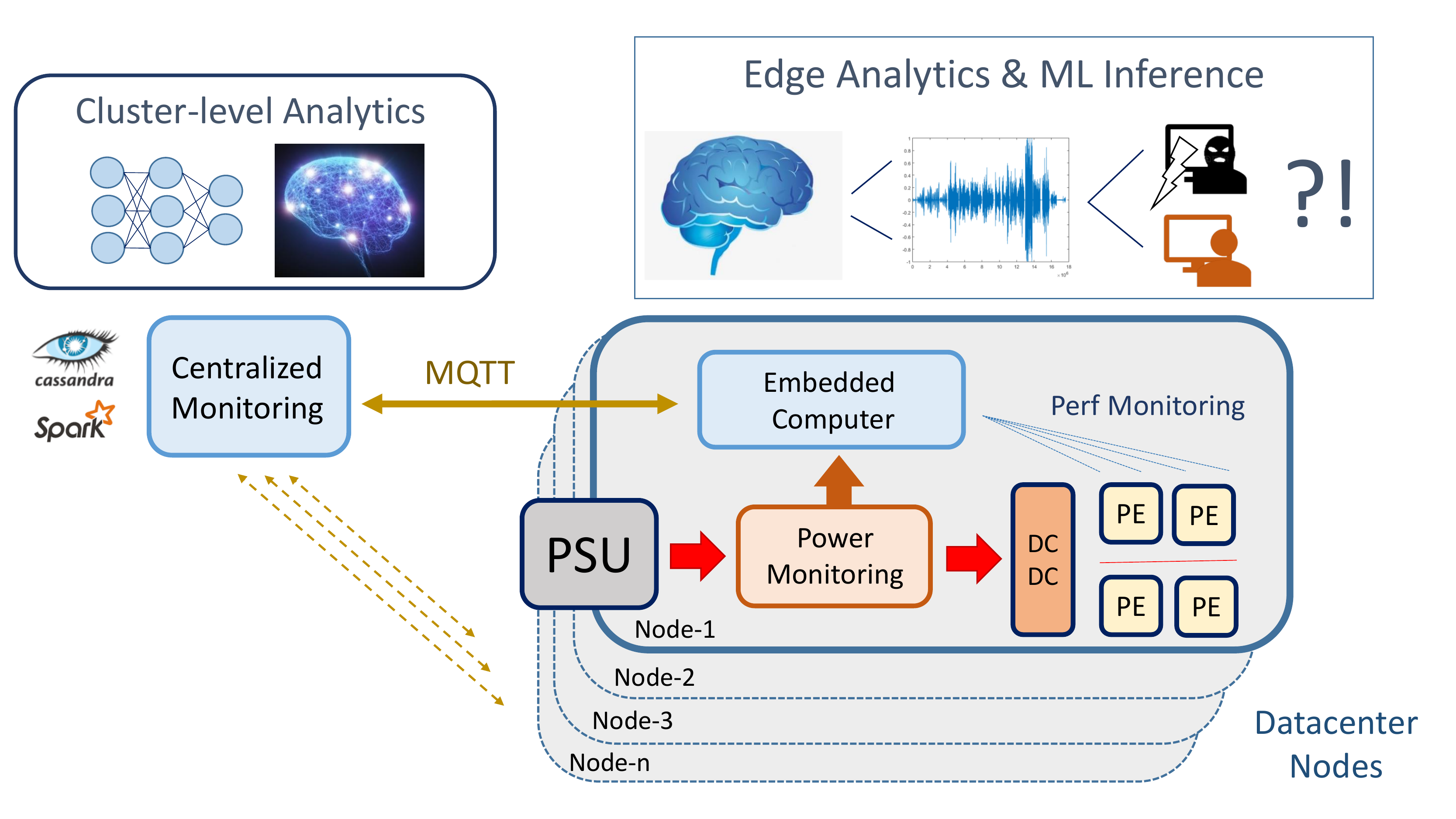}
\caption{Sketch of the \DIG architecture.}
\label{fig:dig}
\end{figure}

\DIG~\cite{DiG_DAAC18} is the first monitoring system for \DCSCs already in production, that allows (i) very high frequency monitoring (\ie up to \SI{20}{\micro\s}), (ii) real-time on-board processing for AI on the edge of the cluster (\eg feature extraction algorithms and ML inference), (iii) ultra-precise time stamping (\ie sub-microseconds synchronizations - below the sampling period - between the cluster nodes) and (iv) vertical integration in a scalable distributed data analytics framework (\ie the open source ExaMon system~\cite{ExaMon}). The monitoring infrastructure is completely out-of-band, scalable, and low cost. Moreover, \DIG is technology agnostic - it is already installed on several HPC systems, based on different CPU architectures (\ie ARM, Intel, and IBM).

As depicted in Figure~\ref{fig:dig}, the monitoring system architecture involves:

\begin{itemize}
    \item a power sensor, which is placed on each \DC compute node, between the power supply unit and the DC-DC (Direct Current to Direct Current) converters that provide power for all the processing elements (PE)\,/\,electrical components within the node. The power sensing is based on a current mirror and shunt resistor to measure the current, and on a voltage divider to measure the voltage. We use it to acquire SoA high-resolution power measurements, with a time granularity of \SI{20}{\micro\s} and a precision below \SI{1}{\percent} ($\sigma$);
    \item an IoT embedded computer (\ie BeagleBone Black - BBB) dedicated for each compute node, that we use to carry out server-level analytics. The system is interfaced with the power sensor via a 12-bit 8-channels SAR ADC, and with existing in-band\,/\,out-of-band telemetries to collect hardware performance counters (\eg Amester~\cite{rosedahl2017}, IPMI~\cite{IPMI}, and RAPL~\cite{RAPL_intel}). Moreover, it includes (i) hardware support for the Precision Time Protocol (PTP), which allows sub-microsecond measurements synchronization~\cite{my_hpcs2016,myAndare18}, (ii) two Programmable Real-Time Units (PRU0 and PRU1), that we exploit for real-time feature extraction on-board, and (iii) an ARM Cortex-A8 processor with NEON technology, useful for DSP processing and edge ML inference (\eg by leveraging the ARM NN SDK\cite{ArmNN}, which enables efficient translation of existing NN frameworks - such as TensorFlow - to ARM Cortex-A CPUs);
    \item a scalable distributed data analytics framework, namely ExaMon~\cite{ExaMon,BartPav19}, that we use to collect at a lower rate - from seconds to milliseconds - power and performance activity of the all cluster and thus to carry out cluster-level analytics. To send data from the distributed monitoring agents (\ie daemons running on the BBBs) to the centralized monitoring unit, we adopted MQTT~\cite{mqtt_ibm}, which is a robust, lightweight and scalable publish-subscribe protocol, already used for large-scale systems both in industry and academia (\eg Amazon, Facebook, \cite{mqtt_ibm}, \cite{mqtt_beneventi}). Moreover, we use Apache Cassandra~\cite{cassandra} to store data in a scalable database and exploit Apache Spark~\cite{spark2016} as Big Data engine to perform cluster-level ML analytics both in streaming and batch mode.
\end{itemize}
\smallskip
\DIG provides a flexible platform to collect fine-grain measurements of the compute nodes activity, and prototype novel anomaly detection methods. Furthermore, \DIG allows to test new strategies for data-science on-board, to handle both the considerable amount of data that otherwise would impact on the communication network and the real-time processing.

\section{\textit{\pae}: Algorithm \& Implementation}\label{sec:pAE}

In this section, we explain the different phases of the \pae approach, namely (i) real-time Feature Extraction and (ii) on-board MD inference, and their implementation in the \DIG infrastructure. Furthermore, we propose a methodology for online training in a \DC.

\subsection{Edge Real-Time Feature Extraction}

For the Feature Extraction phase, we use the PSD with the Welch method~\cite{welch1967}. This technique is based on the periodogram method, with the difference that the estimated power spectrum contains less noise, but with a penalty in the frequency resolution. In particular, the signal is split up into overlapping segments, which are then windowed and used to compute periodograms with the Fast Fourier Transform (FFT). Finally, it is computed the squared magnitude of the result, and the individual periodograms are averaged. We selected it over other approaches - \eg Discrete Wavelet Transform (DWT) or Wavelet Packet Transform (WPT) - for (i) its capability to unveiling relevant frequency components in all the signal bandwidth (\wrt DWT and WPT, which instead provide temporal information at a price of a lower frequency resolution) and (ii) for its low computational complexity~\cite{welch1967}. Moreover, PSDs already proved to be a robust feature extraction method to be used together with NNs, in applications such as speech recognition~\cite{deng2010,wan2017google}.

\begin{figure}[ht]
\centering
\includegraphics[width=3.4in]{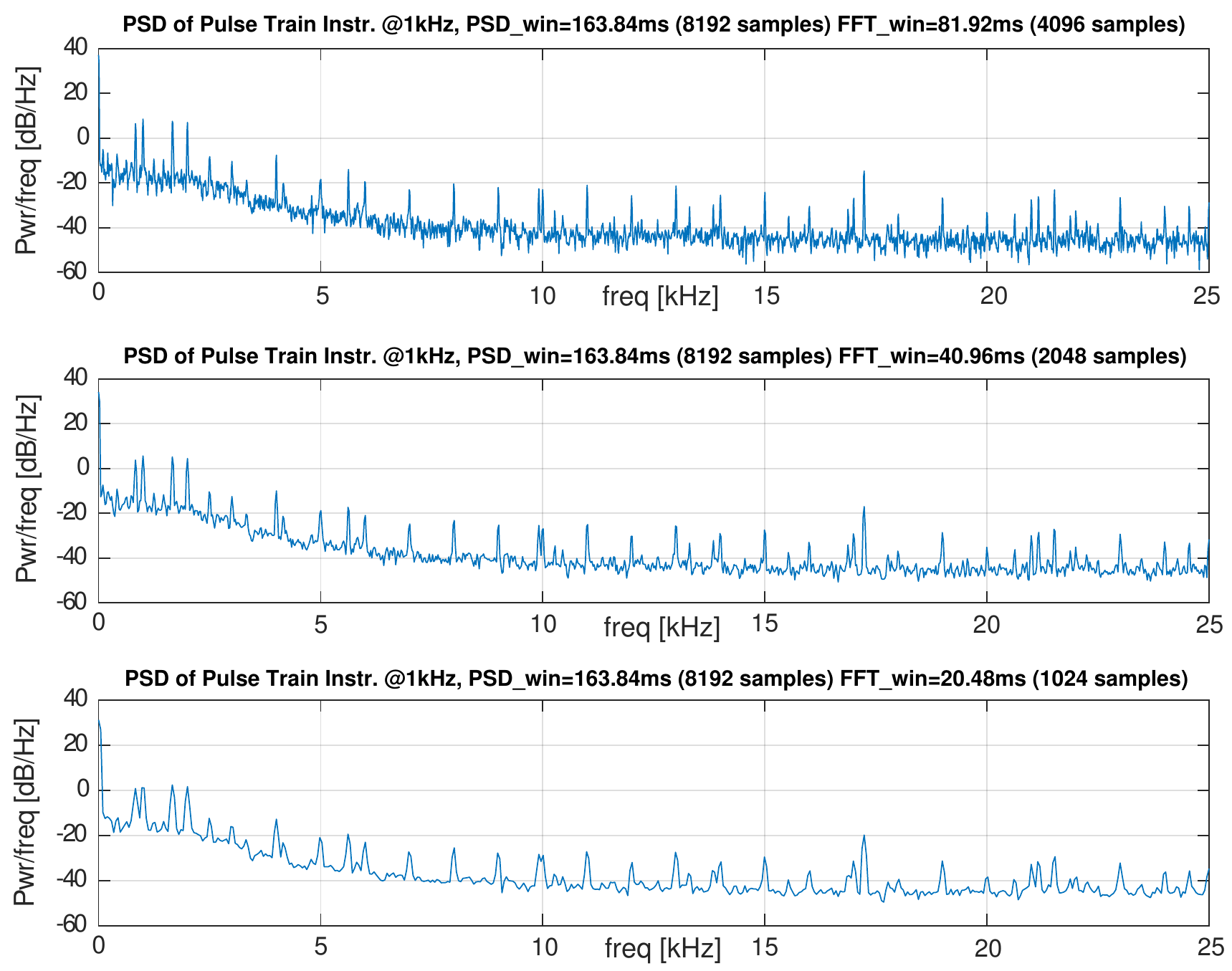}
\caption{Comparison between PSDs of a Pulse Train of instructions at \SI{1}{kHz} for different FFT lengths.}
\label{fig:psd1kHz}
\end{figure}

To find a good time window to use for the FFTs, we run on the monitored compute node a synthetic benchmark that generates a known pattern, and we monitored it with \DIG. In particular, we use a pulse train of instructions at \SI{1}{\kilo\Hz}, where we alternate high load computational phases with idle phases. Figure~\ref{fig:psd1kHz} shows the results of the PSD analysis in a time window of \texttildelow\SI{164}{\milli\s} (\ie 8192 data samples), and different lengths of the FFTs, namely 4096, 2048 and 1024 points (\ie \texttildelow\SI{82}{\milli\s}, \texttildelow\SI{41}{\milli\s} and \texttildelow\SI{20.5}{\milli\s}, respectively). Except for the one at 1024 samples, the other two can detect with an approximately good precision the main peak at \SI{1}{\kilo\Hz} (magnitude greater than \SI{5}{\dB}), plus its harmonics in all the signal bandwidth \SIrange[range-phrase = --]{0}{25}{\kilo\Hz}.

To respect the real-time constraints in the \DIG IoT devices, we selected the PSD with FFTs at 2048 samples, which correspond to an output of 1025 data samples - \ie $(FFT\_samples/2)+1$ - that we use as input features for the phase of MD inference. Finally, we compute consecutive PSDs with sliding windows of \SI{20}{\milli\s} (1000 power samples). As shown in Section~\ref{sec:res}, this decision proved to be a good choice in terms of both (i) MD accuracy and (ii) to create a consistent dataset to compare with SoA MD techniques. However, future works can investigate the results of the MD inference also with different lengths of the PSD, FFTs, and sliding windows, to deploy \pae in other kinds of IoT devices, with different memory\,/\,processing requirements. 

\subsection{Edge Malware Detection Inference}\label{subsec:pAE_inference}

For the ML inference phase, we use an AE, which is a particular kind of Neural Network suitable for Anomaly Detection~\cite{BorghesiAD2019,Wang2019}. As described in Section~\ref{sec:malw}, the idea is to train a model on \quotes{healthy} activity of the \DC (\ie with no malware involved), and try to identify possible anomalies in these signatures when a malware is running in background. With an AE we can do exactly this. By learning to replicate the most salient features in the training data, the AE tries to reconstruct its input $x$ on its output $y$. When facing anomalies, the model it worsens its reconstruction performance. 

In particular, the AE consists of one or more internal (hidden) layers $H$, and involves mainly 2 phases: an encoding phase $H = enc(x)$, and a decoding phase that tries to reconstruct the input $y = dec(H)$ under certain constraints (\eg the dimension of the hidden layers can be smaller than the dimension of the input). These constraints allow to not just simply learn how to reconstruct a perfect copy of the input, namely the identity function $dec(enc(x)) = x$, but instead to learn an approximation to the identity function, so as to output $\hat{x}$ that is similar to $x$, and thus to discover interesting structures about the data. The difference between input and output is the reconstruction error $E_r$, and we can use it as an anomaly score to detect malware. Indeed, after training, the AE can reconstruct healthy data very well (\ie $E_r$ is small and comparable to the error at training time), while failing to do so with anomalous signatures that the AE has not encountered yet (\ie large $E_r$ \wrt the training error).

After an empirical evaluation, we chose a network with 4 fully-connected layers (3 hidden layers plus the output layer), where we implement the phases of encoding-encoding-decoding-decoding, respectively. As shown in our results, this network proved to be a good option in terms of accuracy and computational demands, especially for training and inference time. In particular, the 3 hidden layers employ \{8,4,4\} neurons, while - as in every AE - the output layer has the same dimension of the input, that in our case corresponds to 1025 features (\ie PSD dimension). As regularization term, we employ the L1-norm~\cite{goodfellow2016deep}, while as activation functions for the neurons we use the hyperbolic tangent (tanh) and the rectified linear unit (relu), in the sequence \{tanh-relu-relu-tanh\} for the 4 layers, respectively. Moreover, before processing the PSDs as input features, we pre-process them, by removing the mean and scaling to unit variance.

Lastly, we adopted a threshold-based method, with 2 thresholds, to distinguish between malware and healthy states. The first threshold, namely $T_E$, is related to the reconstruction error and allows us to understand if a PSD is an outlier. To compute $E_r$, we use the Mean Square Error (MSE) between input and output of the AE. Then we tag as anomalies all the PSDs where $E_r$ is greater than $T_E$. The second threshold, $T_O$, is related to the percentage of outliers detected. If this percentage is greater than $T_O$, we detected a potential malware running in the server, and thus we can raise an alarm for the system admin of the \DC.

\subsection{Algorithm Implementation in the \DIG IoT Devices}

\begin{figure}[ht]
\centering
\includegraphics[width=3.4in]{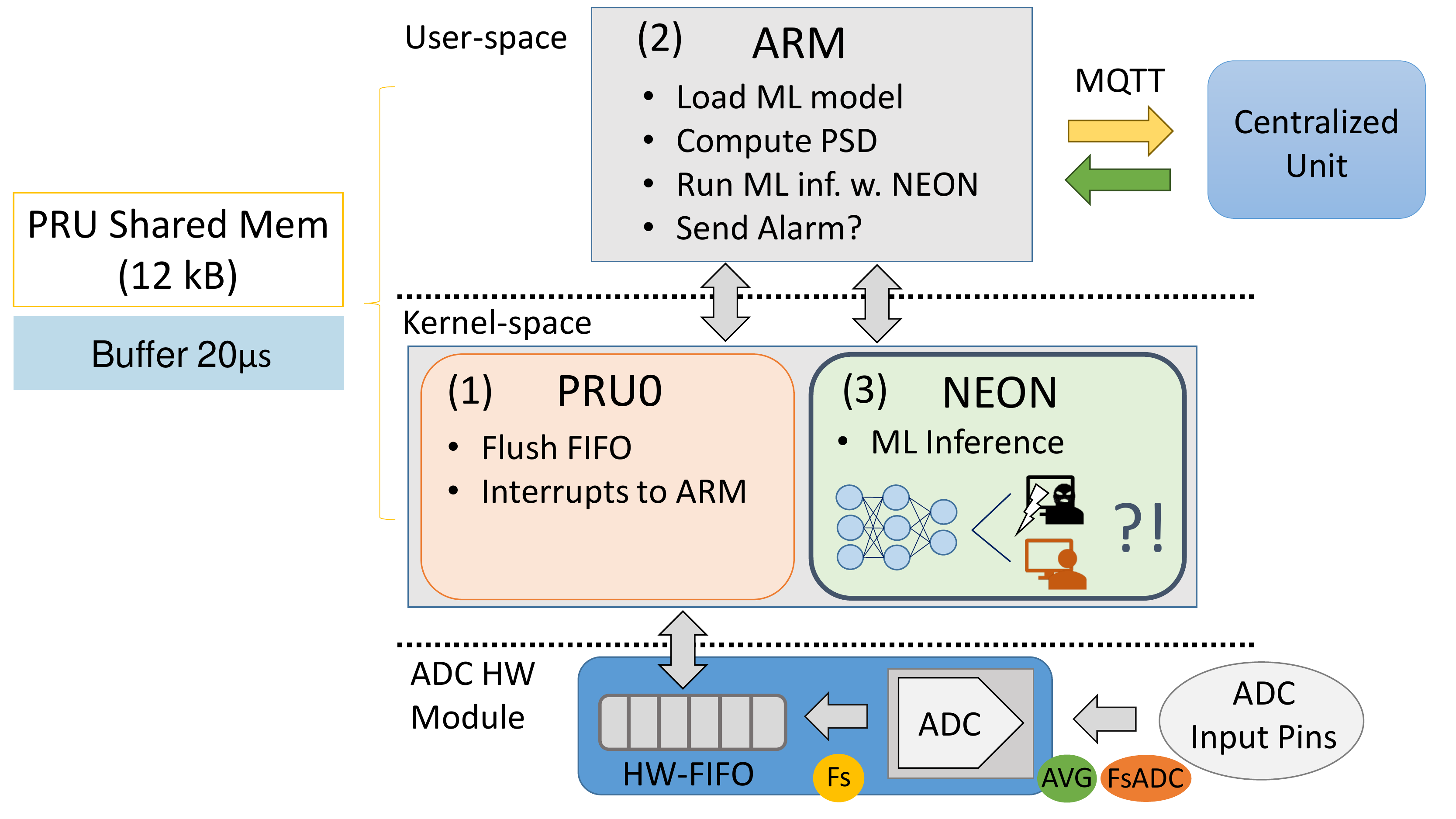}
\caption{Implementation of \pae in the IoT embedded systems of \DIG.}
\label{fig:pru}
\end{figure}

The on-board implementation of the \pae algorithm involves mainly 3 phases: (i) real-time acquisition of the power consumption measurements, that we carry out with the PRU, (ii) real-time computation of the PSD, that we perform in floating-point on the ARM processor, and (iii) real-time MD inference, where we exploit the NEON SIMD architecture extension. Figure~\ref{fig:pru} highlights their implementation in the BBB.

\textit{Phase1:} The bottom layer represents the ADC hardware module. We exploit the ADC continuous sampling mode to continuously sample the two input channels (\ie current and voltage), and store them in a hardware FIFO (HW-FIFO). By tuning the ADC sampling frequency (FsADC) and the hardware averaging (AVG), it is possible to set the frequency (Fs) at which samples enter the software layers. The best trade-off corresponds to a sampling rate (FsADC) of \SI{800}{\kilo S/\s} per channel - maximum rate when using two channels - and hardware average every 16 samples. This is equivalent to \SI{50}{\kilo S/\s} (\ie Fs equals to \SI{20}{\micro \s}) and allows to (i) cover the entire signal bandwidth of the server power consumption, and (ii) reach a measurement precision below \SI{1}{\percent} ($\sigma$) of uncertainty (\aka oversampling and averaging method~\cite{BitRes}). When the HW-FIFO reaches a watermark on the number of samples acquired (we set it to 32), an interrupt is raised and the samples are flushed by the PRU0 into the PRU Shared memory (\SI{12}{\kilo\byte}).

\textit{Phase2:} When a given number of samples is collected (we set it to 2048 samples, which correspond to \SI{4}{\kilo\byte} in memory - \SI{2}{\byte} each), the PRU0 raises an interrupt to the ARM through a message passing protocol named RPMsg. The ARM can then access the PRU Shared Memory to read the samples and store them in RAM. As soon as it reaches a time window of 8192 samples (\SI{163.84}{\milli\s}) it uses the samples to compute the PSD. In this way, we can achieve \SI{100}{\percent} hard real-time demand by not losing power samples, and completely offload the data acquisition phase to the PRU, making the ARM processor available for other tasks, such as computing the PSD and carrying out ML inference. Moreover, by computing the PSD in the ARM processor, we can exploit the Floating Point Unit (which is not available in the PRU) to avoid losing precision when computing the PSD. 

\textit{Phase3:} When the PSD is calculated, the ARM stores it in RAM. After the collection of a batch of PSDs, we exploit the NEON SIMD architecture extension of the ARM to run MD inference. In particular, depending on the activity is currently running in the \DC (\eg system in idle, or application X), the centralized monitoring unit - which has the complete knowledge of the status of the cluster - communicate to the ARM which ML model to load for inference. At this point, if a malware is detected, we send an alarm to the system admins of the \DC.

\subsection{Methodology for Online Training}

For the training phase, we propose a methodology that is suitable to run online on \DCSCs in production, which integrate a \DIG-like monitoring infrastructure (\eg \DAVIDE). It is noteworthy that this approach can be used more in general also for AD, in cases where high-resolution analysis plays a key role in identifying anomalous patterns. In particular, we take acquisitions during healthy-states of the \DC. We compute for these acquisitions the PSDs on \DIG, and pass these data to the corresponding compute nodes for training the AE models (\ie each node trains its AE - for example with GPUs).

This distributed approach allows to (i) reduce both data pre-processing and training time, along with the amount of data to be communicated on the network if using instead a centralized method (\eg pre-processing and training on the centralized monitoring unit); (ii) scale to large \DCSCs. Moreover, the training time and overhead are not a critical concern for the servers, since (i) the training phase takes place only at the beginning, and then at a very low rate, and mostly depending on new applications - or substantially modified versions of previously trained applications - that are running; and (ii) this activity can be scheduled during maintenance periods.

After some experiments, we use for training in our tests the Adagrad algorithm~\cite{adagrad}, with MSE as loss function, a batch size equal to 8, and 5 epochs. After the training phase, the ML models are loaded on the embedded monitoring boards, ready for inference on new data. Notice that each AE is trained with data coming from its corresponding node. This allows to take into account specific activity related to that node, such as particular programs installed and running in background, but also possible differences in the hardware (\eg large variations of material properties when moving to \SIrange[range-phrase = --]{10}{7}{\nano\m} chips~\cite{kachris18hw}).

\section{Experimental Evaluation}\label{sec:res}

In this chapter, we report the results of the experimental evaluation carried out with our approach to detect malware - and anomalies - in a \DCSC, and a comparison with SoA techniques that exploit performance counters. We conclude the section by reporting on the performance of the \pae algorithm running in the \DIG infrastructure.

\subsection{Building Dataset for Analysis and Comparison with SoA}

To create a robust dataset of malware, for testing our approach and compare it with SoA techniques, we downloaded a collection of 95 malware from \textit{virusshare.com}, which is a repository of malware samples available for security research. The collected malware are of any kind, spreading from different type of backdoors, trojans, and ransomware. To emulate a normal activity of the cluster, we acquired signatures of 7 benchmarks, namely the system in idle, plus 6 scientific applications which are widely used for parallel computing, such as Quantum Espresso (QE)~\cite{QE}, HPLinpack (HPL)~\cite{HPL}, High Performance Conjugate Gradients (HPCG)~\cite{HPCG}, Gromacs~\cite{Gromacs} and NAS Parallel Benchmarks (NPB)~\cite{NAS}, that we use in 2 versions, the one that exploits 9 cores (NPB\_btC9) and the one that uses 16 cores (NPB\_btC16). 

For safety reasons, we run the tests on a compute node that is completely isolated from the rest of the network available to the users of our facilities. The node is an Intel Xeon E5-2600 v3 (Haswell). To compare \pae with SoA techniques, we collected for each acquisition both (i) hardware performance counters and (ii) PSDs computed from the high-resolution power measurements. In particular, we analyzed the performance counters suggested by Tang et al.~\cite{tang2014}, plus further metrics available with our open source monitoring tool, ExaMon~\cite{ExaMon}. We summarize a list of the selected metrics in Table~\ref{tab:perf_core} (per-core metrics) and Table~\ref{tab:perf_cpu} (per-cpu metrics). Each table shows the \quotes{raw} metrics, which are the ones we can directly select and monitor, and \quotes{derived} metrics, which are the ones we compute with ExaMon on top of the raw metrics (\eg per-core load, IPS, DRAM energy). As reported in Section~\ref{sec:malw}, the PMU of Haswell Intel processors allows to sample a maximum of only 8 performance counters simultaneously. For this reason, we collected all the listed performance counters in multiple runs of the benchmarks\,/\,malware.

\begin{table}[ht]
\centering
\begin{threeparttable}
\caption{Selected Per-Core Perf Counters Basing on SoA~\cite{tang2014}}
\label{tab:perf_core}
\begin{tabular}{l}
\textbf{Raw Metrics} \\ \noalign{\smallskip}\hline\noalign{\smallskip} 
UOPS\_RETIRED (Retired Micro-ops) \\
ICACHE (Instruction Cache misses) \\
LONGEST\_LAT\_CACHE (Core cacheable demand req. missed L3) \\
MEM\_LOAD\_UOPS\_L3\_HIT\_RETIRED (Retired load uops in L3) \\
BR\_MISP\_RETIRED (Mispredicted branch instr. retired) \\
UOPS\_ISSUED (Uops issued to Reservation Station - RS) \\
IDQ\_UOPS\_NOT\_DELIVERED (Uops not deliv. to RAT per thread) \\
INT\_MISC (Core Cycles the allocator was stalled) \\
BR\_INST\_RETIRED.NEAR\_RETURN (Instr. retired direct near calls) \\
BR\_INST\_RETIRED.NEAR\_CALL (Direct and indirect near call ret.) \\
BR\_INST\_EXEC.ALL\_DIRECT\_NEAR\_CALL (Retired direct calls) \\
BR\_INST\_EXEC.TAKEN\_INDIRECT\_NEAR\_CALL (Ret. indir. calls) \\
DTLB\_STORE\_MISSES.STLB\_HIT (Store operations miss first TLB) \\
ARITH.DIVIDER\_UOPS (Any uop executed by the Divider) \\
DTLB\_LOAD\_MISSES.STLB\_HIT (Load operations miss first DTLB) \\
MEM\_LOAD\_UOPS\_RETIRED.L3\_MISS (Miss in last-level L3) \\
MEM\_LOAD\_UOPS\_RETIRED.L2\_MISS (Miss in mid-level L2) \\
MEM\_LOAD\_UOPS\_RETIRED.L1\_MISS (Ret. load uops miss in L1) \\
BR\_MISP\_EXEC.ALL\_BRANCHES (Retired mispr conditional branch) \\
BR\_MISP\_EXEC.TAKEN\_RETURN\_NEAR (Retired mispr. indir. br.) \\
C3 (Clock cycles in C3 state) \\
C3res (C3 residency - Clock cycl. in C3 state between 2 sampling time) \\
C6 (Clock cycles in C6 state) \\
C6res (C6 residency - Clock cycl. in C6 state between 2 sampling time) \\
temp (Cores temperature) \\ \noalign{\smallskip}\hline\noalign{\smallskip} \\
\textbf{Derived Metrics} \\ \noalign{\smallskip}\hline\noalign{\smallskip}
load\_core (per-core load) \\
IPS (Instructions per Second) \\
CPI (Cycles per Instruction) \\ \noalign{\smallskip}\hline\noalign{\smallskip}
\end{tabular}
\begin{tablenotes}
  \small 
  \item \textit{\textbf{Note.} More info about the metrics are available in~\cite{ExaMon,intel_perf,tang2014}.}
\end{tablenotes}
\end{threeparttable}
\end{table}

\begin{table}[ht]
\centering
\begin{threeparttable}
\caption{Selected Per-CPU Perf Counters}
\label{tab:perf_cpu}
\begin{tabular}{l}
\textbf{Raw Metrics} \\ \noalign{\smallskip}\hline\noalign{\smallskip} 
C2 (Clock cycles in C2 state) \\
C3 (Clock cycles in C3 state) \\
C6 (Clock cycles in C6 state) \\
C2res (C2 residency) \\
C3res (C3 residency) \\
C6res (C6 residency) \\
freq\_ref (Core frequency) \\
erg\_units (Energy units) \\
temp\_pkg (Package temperature) \\ \noalign{\smallskip}\hline\noalign{\smallskip} \\
\textbf{Derived Metrics} \\ \noalign{\smallskip}\hline\noalign{\smallskip}
erg\_dram (Energy DRAM consumed) \\
erg\_pkg (Package Energy consumed) \\  \noalign{\smallskip}\hline\noalign{\smallskip}
\end{tabular}
\begin{tablenotes}
  \small 
  \item \textit{\textbf{Note.} More info in~\cite{ExaMon,intel_perf}.}
\end{tablenotes}
\end{threeparttable}
\end{table}

For the performance metrics, we use a time granularity of \SI{20}{\milli\s} \ie in line with SoA \DCSCs monitoring tools~\cite{BartPav19}), dedicating 2 cores for ExaMon to prevent that the in-band monitoring can affect the measurements with noise. For this purpose, we exploit \textit{isolcpus}, which is a kernel parameter that can be set from the boot loader configurations. Instead, for the high-resolution power measurements, we can benefit of a time granularity of \SI{20}{\micro\s}, thanks to the out-of-band monitoring of \DIG that allows to do not subtract computing resources from users. In particular, we compute the PSDs in time windows of \SI{163.84}{\milli\s} (8192 samples), with FFT of 2048 points, and sliding windows of 1000 samples between two consecutive PSDs. Notice that the sliding window corresponds to \SI{20}{\milli\s}, which is consistent with the granularity of the performance counters. In this way, we can construct a dataset that has for each row the different samples acquired over time, and for each column the several features (\ie performance counters and PSDs). Moreover, to show that simple coarse grain statistics of the power measurements are not enough to detect malware, we include as features also standard deviation, mean, max, and min value in the interval of \SI{20}{\milli\s}.

To build the dataset, we acquired first signatures of healthy benchmarks (normal activity of the \DC with no malware involved), and then signatures of the same benchmarks but with malware in background. In particular, we collected 30 acquisitions for each benchmark and 2 acquisitions for each of the 95 malware running in background with every benchmark. Notice that for each run of the malware, we use a completely fresh installation of the OS, to prevent that different malicious activities can interfere and thus invalidate the analysis. We release both dataset and software to analyze the data open source, at the following link~\cite{pae_dataset}.

\subsection{Malware Detection Results}

To evaluate the performance of our approach, and compare it with SoA techniques for MD, we tested the acquisitions in our dataset with both performance metrics and PSD features, and with different ML algorithms suitable for Anomaly and Malware Detection, namely oc-SVM~\cite{tang2014}, Isolation Forest (IF)~\cite{IF} and Autoencoders. For oc-SVM and IF, we use the Scikit-learn~\cite{scikit_learn} implementation developed in python, while for the AE we use Keras~\cite{keras} with TensorFlow~\cite{TF} as back-end. 

After an empirical evaluation, we adopted standardization (Scikit-learn \textit{StandardScaler}) like in \pae (settings described in Section~\ref{subsec:pAE_inference}), to pre-process the features also for oc-SVM and IF. Moreover, for the oc-SVM we use the Principal Component Analysis (PCA) to reduce the dimension of the feature space to 25 components (\aka dimensionality reduction), and the polynomial kernel with 0.1 as kernel coefficient (\ie \quotes{gamma} in the Scikit-learn API). Instead, for IF we obtain better results without PCA, and with the contamination parameter set to \quotes{auto} (\ie decision function threshold determined as in~\cite{IF}). Finally, as in \pae, we use for oc-SVM and IF a threshold-based method, but in their case with only one threshold on the percentage of outliers found in the benchmark, that we set to \SI{30}{\percent} (\ie if the number of anomalous samples is above this threshold, we label it as a malware).

To train the ML models (1 model for each benchmark), we use only healthy benchmarks, while for validation (useful to find proper settings for our models without incurring to overfitting) and test, we use both healthy and malicious signatures. In particular, we split the healthy benchmarks subset to \SI{60}{\percent} for training, \SI{20}{\percent} for validation and \SI{20}{\percent} for test, while we split the malicious subset to \SI{50}{\percent} for validation and \SI{50}{\percent} for test. Notice also that together with the analysis of the performance counters, we include the coarse-grain power statistics at \SI{20}{\milli\s}.

\begin{table*}[ht!]
\centering
\begin{threeparttable}
\caption{Final results of the MD analysis and comparison with SoA approaches}
\label{tab:results}
\begin{tabular}{clllllll}
& \multicolumn{1}{c}{} & \multicolumn{3}{c}{Perf Metrics + Power Statistics @20ms} & \multicolumn{3}{c}{\textbf{PSD of Power @20us}} \\ \noalign{\smallskip}\cline{3-8}\noalign{\smallskip}
ML Approach & Benchmark & \multicolumn{1}{c}{False Alarm} & \multicolumn{1}{c}{Malware Miss} & \multicolumn{1}{c}{\fscore} & \multicolumn{1}{c}{False Alarm} & \multicolumn{1}{c}{Malware Miss} & \multicolumn{1}{c}{\fscore} \\ \noalign{\smallskip}\hline\noalign{\smallskip}
& Idle & \multicolumn{1}{c}{1.000} & \multicolumn{1}{c}{0.042} & \multicolumn{1}{c}{0.645} & \multicolumn{1}{c}{1.000} & \multicolumn{1}{c}{0.000} & \multicolumn{1}{c}{0.664} \\
& QE & \multicolumn{1}{c}{1.000} & \multicolumn{1}{c}{0.011} & \multicolumn{1}{c}{0.660} & \multicolumn{1}{c}{0.000} & \multicolumn{1}{c}{0.221} & \multicolumn{1}{c}{0.876} \\
& HPL & \multicolumn{1}{c}{1.000} & \multicolumn{1}{c}{0.095} & \multicolumn{1}{c}{0.621} & \multicolumn{1}{c}{0.167} & \multicolumn{1}{c}{0.968} & \multicolumn{1}{c}{0.053} \\
oc-SVM & HPCG & \multicolumn{1}{c}{1.000} & \multicolumn{1}{c}{0.600} & \multicolumn{1}{c}{0.332} & \multicolumn{1}{c}{0.167} & \multicolumn{1}{c}{0.032} & \multicolumn{1}{c}{0.906} \\
& Gromacs & \multicolumn{1}{c}{1.000} & \multicolumn{1}{c}{0.074} & \multicolumn{1}{c}{0.631} & \multicolumn{1}{c}{0.000} & \multicolumn{1}{c}{0.032} & \multicolumn{1}{c}{0.984} \\
& NPB\_btC9 & \multicolumn{1}{c}{1.000} & \multicolumn{1}{c}{0.853} & \multicolumn{1}{c}{0.137} & \multicolumn{1}{c}{0.000} & \multicolumn{1}{c}{1.000} & \multicolumn{1}{c}{0.000} \\
& NPB\_btC16 & \multicolumn{1}{c}{1.000} & \multicolumn{1}{c}{0.884} & \multicolumn{1}{c}{0.109} & \multicolumn{1}{c}{0.000} & \multicolumn{1}{c}{0.042} & \multicolumn{1}{c}{0.978} \\ \noalign{\smallskip}\cline{2-8}\noalign{\smallskip}
& Overall & \multicolumn{1}{c}{1.000} & \multicolumn{1}{c}{0.365} & \multicolumn{1}{c}{0.480} & \multicolumn{1}{c}{0.190} & \multicolumn{1}{c}{0.328} & \multicolumn{1}{c}{0.721} \\ \noalign{\smallskip}\hline\noalign{\smallskip}
& Idle & \multicolumn{1}{c}{0.000} & \multicolumn{1}{c}{0.053} & \multicolumn{1}{c}{0.973} & \multicolumn{1}{c}{0.000} & \multicolumn{1}{c}{0.979} & \multicolumn{1}{c}{0.041} \\
& QE & \multicolumn{1}{c}{0.000} & \multicolumn{1}{c}{0.137} & \multicolumn{1}{c}{0.927} & \multicolumn{1}{c}{0.000} & \multicolumn{1}{c}{0.968} & \multicolumn{1}{c}{0.061} \\
& HPL & \multicolumn{1}{c}{0.000} & \multicolumn{1}{c}{0.642} & \multicolumn{1}{c}{0.527} & \multicolumn{1}{c}{0.000} & \multicolumn{1}{c}{0.979} & \multicolumn{1}{c}{0.041} \\
IF & HPCG & \multicolumn{1}{c}{0.000} & \multicolumn{1}{c}{0.000} & \multicolumn{1}{c}{1.000} & \multicolumn{1}{c}{0.000} & \multicolumn{1}{c}{1.000} & \multicolumn{1}{c}{0.000} \\
& Gromacs & \multicolumn{1}{c}{0.000} & \multicolumn{1}{c}{0.874} & \multicolumn{1}{c}{0.224} & \multicolumn{1}{c}{0.000} & \multicolumn{1}{c}{1.000} & \multicolumn{1}{c}{0.000} \\
& NPB\_btC9 & \multicolumn{1}{c}{1.000} & \multicolumn{1}{c}{0.000} & \multicolumn{1}{c}{0.664} & \multicolumn{1}{c}{0.000} & \multicolumn{1}{c}{1.000} & \multicolumn{1}{c}{0.000} \\
& NPB\_btC16 & \multicolumn{1}{c}{0.667} & \multicolumn{1}{c}{0.000} & \multicolumn{1}{c}{0.748} & \multicolumn{1}{c}{0.000} & \multicolumn{1}{c}{0.958} & \multicolumn{1}{c}{0.081} \\ \noalign{\smallskip}\cline{2-8}\noalign{\smallskip}
& Overall & \multicolumn{1}{c}{0.238} & \multicolumn{1}{c}{0.244} & \multicolumn{1}{c}{0.758} & \multicolumn{1}{c}{0.000} & \multicolumn{1}{c}{0.983} & \multicolumn{1}{c}{0.033} \\ \noalign{\smallskip}\hline\noalign{\smallskip}
& Idle & \multicolumn{1}{c}{0.000} & \multicolumn{1}{c}{0.579} & \multicolumn{1}{c}{0.593} & \multicolumn{1}{c}{\textbf{0.000}} & \multicolumn{1}{c}{\textbf{0.032}} & \multicolumn{1}{c}{\textbf{0.984}} \\
& QE & \multicolumn{1}{c}{0.000} & \multicolumn{1}{c}{0.937} & \multicolumn{1}{c}{0.119} & \multicolumn{1}{c}{\textbf{0.000}} & \multicolumn{1}{c}{\textbf{0.000}} & \multicolumn{1}{c}{\textbf{1.000}} \\
& HPL & \multicolumn{1}{c}{0.000} & \multicolumn{1}{c}{0.947} & \multicolumn{1}{c}{0.100} & \multicolumn{1}{c}{\textbf{0.000}} & \multicolumn{1}{c}{\textbf{0.000}} & \multicolumn{1}{c}{\textbf{1.000}} \\
\textbf{AE} & HPCG & \multicolumn{1}{c}{0.167} & \multicolumn{1}{c}{0.042} & \multicolumn{1}{c}{0.901} & \multicolumn{1}{c}{\textbf{0.000}} & \multicolumn{1}{c}{\textbf{0.000}} & \multicolumn{1}{c}{\textbf{1.000}} \\
& Gromacs & \multicolumn{1}{c}{0.000} & \multicolumn{1}{c}{0.947} & \multicolumn{1}{c}{0.100} & \multicolumn{1}{c}{\textbf{0.000}} & \multicolumn{1}{c}{\textbf{0.000}} & \multicolumn{1}{c}{\textbf{1.000}} \\
& NPB\_btC9 & \multicolumn{1}{c}{0.000} & \multicolumn{1}{c}{0.084} & \multicolumn{1}{c}{0.956} & \multicolumn{1}{c}{\textbf{0.000}} & \multicolumn{1}{c}{\textbf{0.000}} & \multicolumn{1}{c}{\textbf{1.000}} \\
& NPB\_btC16 & \multicolumn{1}{c}{0.000} & \multicolumn{1}{c}{0.189} & \multicolumn{1}{c}{0.895} & \multicolumn{1}{c}{\textbf{0.000}} & \multicolumn{1}{c}{\textbf{0.000}} & \multicolumn{1}{c}{\textbf{1.000}} \\ \noalign{\smallskip}\cline{2-8}\noalign{\smallskip}
& Overall & \multicolumn{1}{c}{0.024} & \multicolumn{1}{c}{0.532} & \multicolumn{1}{c}{0.627} & \multicolumn{1}{c}{\textbf{0.000}} & \multicolumn{1}{c}{\textbf{0.005}} & \multicolumn{1}{c}{\textbf{0.998}} \\ \noalign{\smallskip}\hline\noalign{\smallskip}
\end{tabular}
\begin{tablenotes}
  \small 
  \item \textit{\textbf{Note.} For best results we want False Alarm rate (healthy benchmarks seen as malware) and Malware Miss rate (malware not detected) close to 0, while F1-score close to 1. Results related to \pae are highlighted in bold.}
\end{tablenotes}
\end{threeparttable}
\end{table*}

Table~\ref{tab:results} reports the results of our analysis in the test set. Namely, the (i) False Alarm (FA) rate (= False Positive - FP - rate) - \ie healthy benchmarks erroneously labeled as malware; (ii) the Malware Miss (MM) rate (= False Negative - FN - rate) - \ie malware not detected; and (iii) the weighted \fscore~\cite{AysePerf2019} (best value at 1, and worst at 0), which measure the test accuracy based on the following formula:

\begin{equation}
F1score = \frac{2TP \cdot W_M}{ 2TP \cdot W_M + FP \cdot W_H + FN \cdot W_M}
\end{equation}

where we weighted the True Positives (TP), FP, and FN by the number of instances of each class (Malware and Healthy), via the two weights W\textsubscript{M} and W\textsubscript{H}, to take into account the imbalance of the dataset between number of malware and number of healthy acquisitions. Indeed, if we do not consider a weighted \fscore, then a naive classifier that marks every sample as malware would achieve an overall \fscore of 0.95, as approximately \SI{95}{\percent} of our dataset consists of malware acquisitions. Moreover, notice that, as we are focusing on the detection of malware, it is really important that the MM rate is close to zero, and of course, also that the FA rate is reasonably low, to avoid too many false alarms to handle, and the \fscore is close to 1 (best accuracy).

Notice that, to the best of our knowledge, this is the first work based on fine grain monitoring of power and performance, that targets \SCs and \DC compute nodes, along with their requirements (\eg scalability, and reasonable overhead for in-band monitoring, to do not impact computing resources), and that reports a comprehensive analysis with a vast number of malware. When comparing with the analysis via performance counters, in line with other SoA works in literature targeting anomaly detection in \SCs~\cite{AysePerf2019,Borghesi19}, we obtain superior results when using IF and AE, rather than oc-SVM. In particular, in our experience the main problem with oc-SVM is that the feature space is not well separable for performance counters, and thus it is difficult to find a good tuning (\eg kernel coefficient, PCA components, etc.) and set a proper threshold to distinguish between normal activity and malware (\ie both healthy and malicious signatures are always seen as anomaly, leading to a FA rate of 1 and an overall F1-score of 0.48).

Instead, when using IF with performance metrics (+ coarse-grain power statistics) we observed a high F1-score, with \SI{0}{\percent} of FA rate, in the system in idle, HPCG, and QE, while we obtained a poor F1-score for Gromacs and HPL, and a really low F1-score with high FA rate for NPB\_btC9, and NPB\_btC16. When looking at the overall F1-score, we obtain a low value of 0.758, which is in line with results in~\cite{das2019sok,Zhou2018}, when using tree-based models with performance counters for malware detection (reason why they do not advice this technique for security purposes). Finally, in the case of AE, we obtain poor scores for all benchmarks, except for HPCG, NPB\_btC9, and NPB\_btC16, and an overall F1-score of 0.627.

To understand which ML algorithm performs better with PSDs of fine-grain power measurements as input features, we tested all previous methods, and report the results in the three rightmost columns of Table~\ref{tab:results}. In particular, when using oc-SVM we can find slightly better outcomes than using the same approach with performance counters, with an overall F1-score od 0.721 and a FA rate of 0.19. In particular, we observed a good F1-score with \SI{0}{\percent} of FA rate for Gromacs and NPB\_btC16 (0.978 and 0.984, respectively), while slightly worse results for QE and HPCG (where the model was not able to detect all healthy benchmarks). We obtained then a poor F1-score with HPL and NPB\_btC9, and even did not find a good threshold for the system in idle (\ie the model always sees everything as malware). In the case of IF, the results drop to really poor performance, with an overall F1-score close to zero as the MM rate is close to 1 (\ie the model struggles to distinguish between the malicious and healthy patterns).

Lastly, the combination AEs + PSDs (\ie \pae method, highlighted in bold in the table) shows promising results. In particular, we found that the reconstruction error is definitely larger during anomalous periods compared to the normal ones, and thus by setting proper thresholds (\eg 0.91 for the reconstruction error, and \SI{30}{\percent} for the quantity of anomalous samples in the benchmarks) we can obtain an F1-score equal to 1 for almost all benchmarks (overall F1-score equal to 0.998), and both a FA and MM rate close to \SI{0}{\percent}. Comparing these scores with previous results, \pae outperforms the best tested SoA method (IF) with an improvement in accuracy of \SI{24}{\percent}. As the \pae approach is completely out-of-band (zero overhead on the computing resources) and simple to implement in off-the-shelf low-cost IoT devices, we believe it can push the boundaries of security in \DCs to new levels.

\subsection{Computational Load \& Scalability}

With respect to the computational load, \pae is a very lightweight approach. Thanks to the PRU offloading, we can run the whole feature extraction phase in the \DIG IoT devices with less than \SI{1}{\percent} of the ARM CPU (used to transfer data from the PRU shared memory into the main RAM). Moreover, we can compute PSDs in the ARM - length of 8129 samples, FFTs of 2048 samples, and sliding window of 1000 samples (\ie \SI{20}{\milli\s}) - within \texttildelow\SI{19}{\milli\s}, respecting the real-time constraint of the PSD sliding window. It is noteworthy that we used not optimized code for the implementation of the PSD (\ie Welch method in python), which means we can further improve this performance. Then, we can run edge ML inference in batch of 500 PSDs (\ie every \SI{10}{\s}, considering a sliding window of \SI{20}{\milli\s}), exploiting the NEON with Keras and TensorFlow, in less than \SI{0.8}{\s}. Future works can study the performance of lightweight frameworks, such as TensorFlow Light~\cite{TF_Light} or ARM NN~\cite{ArmNN}, and also the accuracy of \pae with different length of PSD, FFTs and sliding windows. Finally, the training phase took for us around \SI{36}{\s} on a \DC server equipped with a GeForce GTX 1080 NVIDIA (GPU usage between \SIrange[range-phrase = --]{14}{15}{\percent}), and with code written in Keras plus TensorFlow for GPU as a back-end.

We highlight also that \pae is a highly scalable AI approach, thanks to the edge computing paradigm. As a matter of example, supposing to carrying out the approach without edge computing in a SoA large-scale system, such as Sunway TaihuLight (\texttildelow41 thousand compute nodes~\cite{Fu2016}), we should handle a rate of \texttildelow\SI{20.5}{\giga\byte/\s} (\ie 41~k nodes $\times$ \SI{50}{\kilo S/\s} per node $\times$ \SI{10}{\byte} for each sample, which includes raw power and timestamp). Considering that it would be impossible to analyze this amount of data offline (the database capacity would reach its limit in few minutes), and that for real-time computing with only a centralized monitoring unit several bottlenecks would arise (\eg high-latency, high SW overhead to handle the data, high network traffic burden, and also the possibility to loose samples), edge AI is the right direction to face this challenge. With its design, \pae requires only few megabytes for training and a few bytes to send alarms when potential malware are detected.

Moreover, we underline the fact that using performance metrics, like in SoA techniques, but with a high monitoring rate to detect malware, would be hard in Top~500 clusters, like Sunway TaihuLight. Indeed, just as matter of example, supposing to collect all 25 per-core performance metrics reported in Table~\ref{tab:perf_core} at the granularity of \SI{20}{\micro\s}, and considering that Sunway TaihuLight integrates 260 cores per CPU~\cite{Fu2016}, this would result to a rate of \SI{325}{\mega S/\s} to be handled either in-band in the compute servers with live analysis (with the consequent overhead in the computing resources), or out-of-band by sending these data to a centralized monitoring unit, with the consequent overhead in the network infrastructure (\ie roughly \SI{13.3}{\tera S/\s}).

Finally, an interesting extension of our approach is to use open-source distributed stream processing systems, like Storm, Flink, or Spark Streaming~\cite{IoT_StormBench17,ShmaIoT18,IsahSurvey19} to carry our real-time analytics. As shown in several works in literature (\eg~\cite{ChanthakitIoT19,Syafrudin18,Chintapalli16}), the common use-case exploits a distributed messaging system, like Apache Kafka~\cite{thein2014apache}, that runs on the IoT devices and sends data to dedicated servers hosting the streaming analytics software. These servers carry out the distributed real-time analytics services, with frameworks such as Storm. As an example, as reported in~\cite{storm_website}, Storm can handle a data rate of \SI{1}{\mega S/\s} per node with messages of \SI{100}{\byte} on servers that include 2 Intel Xeon E5645, running at \SI{2.4}{\giga\Hz} and with \SI{24}{\giga\byte} of memory. In other words, at the cost of using additional dedicated servers it is possible to benefit from a robust and flexible approach for training, and also for maintenance, deployment, and update of the streaming service on the edge. Of course, drawbacks in our case are that (i) to do not overwhelm the data network, the processing servers have to be as close as possible to the data source (\ie BBB), and (ii) the cost of the extra servers for distributed processing would grow with the size of the data center. 

Other possibilities would be (i) to run \pae on the \DIG IoT devices as we do, and then use these distributed systems only for training and deployment of the ML models; or (ii) execute these stream services directly on the IoT devices. However, as shown in~\cite{IoT_StormBench17}, which benchmarks them in a Raspberry Pi 3 (ARMv8 Quad-Core \SI{1}{\giga\byte} RAM vs. ARM A8 Single-Core \SI{1}{\giga\byte} RAM in our BBB), the resources needed (memory and CPU load) for running these services are not negligible, even considering only data ingestion and delivery and not considering actual computation. A more detailed quantitative assessment of the feasibility of this approach with our specific edge hardware platform (BBB) is an interesting topic for future work.

\section{Conclusion}\label{sec:end}

This work reports on novel method - namely \pae - to increasing the cybersecurity of \DCSCs, involving AI-powered edge computing. The method targets real-time malware detection running on an out-of-band IoT-based monitoring system for \DCs, and involves feature extraction analysis based on Power Spectral Density of power measurements, along with AE Neural Networks. Results obtained with a robust dataset of malware are promising, outperforming SoA techniques by \SI{24}{\percent} in accuracy, with an overall F1-score close to 1, and a False Alarm and Malware Miss rate close to \SI{0}{\percent}. Moreover, we propose a methodology suitable for online training in \DCSCs in production, and release SW code and dataset open source~\cite{pae_dataset}. We envisage this approach as support (pre-filter) for Intrusion Detection Systems (IDS), and encourage further research in this direction. Finally, the approach can also be used in a more general context of Anomaly Detection, opening new perspectives for future works.



\section*{Acknowledgment}

This work has been partially supported by the EU H2020 project IoTwins (g.a. 857191).

\ifCLASSOPTIONcaptionsoff
  \newpage
\fi

\bibliographystyle{IEEEtran}

\begin{IEEEbiography}[{\includegraphics[width=1in,height=1.25in,clip,keepaspectratio]{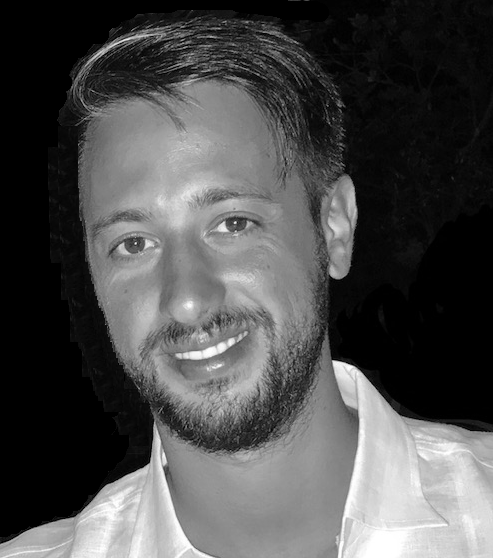}}]{Antonio Libri}
received a M.Sc. degree in electrical engineering from the University of Genova, Italy, in 2013, with a thesis on Wireless Sensor Networks carried out at University College Cork, Ireland. After two years of working as an embedded software engineer in Socowave Ltd, Cork, Ireland, he joined in 2015 ETH Zurich, Switzerland, pursuing a Ph.D. degree. His research interests focus on data monitoring, synchronization, and AI analytics for automation and control of Data Centers\,/\,Supercomputers.
\end{IEEEbiography}

\begin{IEEEbiography}[{\includegraphics[width=1in,height=1.25in,clip,keepaspectratio]{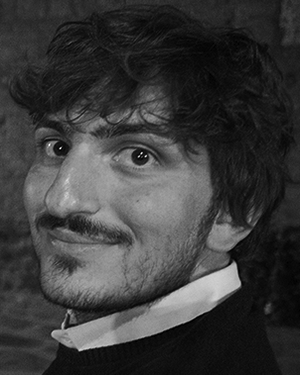}}]{Andrea Bartolini}
received the PhD degree in electrical engineering from University of Bologna, Italy, in 2011. He is currently assistant professor at the Department of Electrical, Electronic and Information Engineering (DEI), University of Bologna. Before, he was post-doctoral researcher in ETH Zurich, Switzerland. His research interests concern dynamic resource management, ranging from embedded to HPC systems with special emphasis on software-level thermal and power-aware techniques.
\end{IEEEbiography}

\begin{IEEEbiography}[{\includegraphics[width=1in,height=1.25in,clip,keepaspectratio]{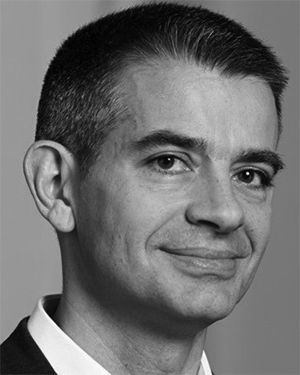}}]{Luca Benini}
is professor of Digital Circuits and Systems at ETH Zurich, Switzerland, and is also professor at University of Bologna, Italy. His research interests are in energy-efficient multicore SoC and system design, smart sensors and sensor networks. He has published more than 1000 papers in peer-reviewed international journals and conferences, four books and several book chapters. He is a fellow of the ACM and Member of the Academia Europea, and is the recipient of the IEEE CAS Mac Van Valkenburg Award 2016.
\end{IEEEbiography}


\vfill


\end{document}